# WikiFactDiff: A Large, Realistic, and Temporally Adaptable Dataset for Atomic Factual Knowledge Update in Causal Language Models

**Hichem Ammar Khodja[1, 2]   Frédéric Béchet[2,3]   Quentin Brabant[1]**
**Alexis Nasr[2]   Gwénolé Lecorvé[1]**

(1) Orange - *Lannion, France*
(2) Aix-Marseille Univ, CNRS, LIS, UMR 7020 - *Marseille, France*
(3) International Laboratory on Learning Systems (ILLS - IRL2020 CNRS)

`hichem.ammarkhodja@orange.com, frederic.bechet@lis-lab.fr, quentin.brabant@orange.com,`
`alexis.nasr@lis-lab.fr, gwenole.lecorve@orange.com`

## Abstract

The factuality of large language model (LLMs) tends to decay over time since events posterior to their training are "unknown" to them. One way to keep models up-to-date could be factual update: the task of inserting, replacing, or removing certain simple (atomic) facts within the model. To study this task, we present WikiFactDiff, a dataset that describes the evolution of factual knowledge between two dates as a collection of simple facts divided into three categories: new, obsolete, and static. We describe several update scenarios arising from various combinations of these three types of basic update. The facts are represented by subject-relation-object triples; indeed, WikiFactDiff was constructed by comparing the state of the Wikidata knowledge base at 4 January 2021 and 27 February 2023. Those fact are accompanied by verbalization templates and cloze tests that enable running update algorithms and their evaluation metrics. Contrary to other datasets, such as zsRE and CounterFact, WikiFactDiff constitutes a realistic update setting that involves various update scenarios, including replacements, archival, and new entity insertions. We also present an evaluation of existing update algorithms on WikiFactDiff.

## 1. Introduction

In addition to reliability issues due to their statistical nature, Large Language Models (LLMs) suffer from a static, time-stopped nature: they only get to learn facts up to the date when their training data was collected (Lazaridou et al., 2021). They can thus propagate outdated information, which can have significant real-world implications, e.g., in domains like healthcare or politics. Therefore, knowing how to update the facts known by these models is crucial to ensure their utility and relevance, as well as the global reliability of all artificial intelligence applications based on them.

While the concept of knowledge is broad (including knowledge about facts, linguistics, procedures, etc.), updating factual knowledge is currently a particularly active area of research. Especially, as opposed to traditional global update approaches through fine-tuning, a recent approach involves atomic updates, i.e., considering only a single fact for the update. In the literature, facts are commonly represented as (subject, relation, object) triples, e.g., (India, head of state, Ram Nath Kovind). In general, numerous scenarios for updating can occur within this framework. For example, a new fact may emerge (e.g., an author publishing a new book), or a previously valid fact may become invalid (e.g., a sports player leaving her/his team). A new fact can also complement an existing one (e.g., an organization gaining a new member) or replace an old one (e.g., a team changing its coach). Moreover, updates can occur not only on existing entities but also with the emergence of new entities (e.g., a new scientific theory). However, to our knowledge, current update algorithms and datasets are limited to the sole scenario of replacement, and on existing entities (Yao et al., 2023). Furthermore, the updates in these works are unrealistic (e.g., changing Albert Einstein's field of expertise from Physics to Biology), which introduces challenges in maintaining global coherence of knowledge (e.g., in this case, would E=MC² also be a result of biology too?). This does not reflect the actual usage of these techniques in real-world applications.

| Subject | Relation | Object | Class |
|---|---|---|---|
| Japan | Population | 125.96M | obsolete |
| | | 125.44M | new |
| Cristiano Ronaldo | member of sports team | Portugal national association football team | static |
| | | Juventus F.C. | obsolete |
| | | Al-Nassr | new |
| USA | head of government | Donald Trump | obsolete |
| | | Joe Biden | new |
| Vyacheslav Geraschenko | coach of sports team | FC Smorgon | obsolete |
| | | FC Dnepr Mogilev | new |
| ChatGPT | instance of | language model | new |
| | | ... | new |
| | inception | 30 November 2022 | new |
| | ... | ... | new |

Table 1: Samples of updates from WikiFactDiff. Every triple whose subject is "ChatGPT" is labeled *new* because this entity is new.

Addressing these limitations, we introduce WikiFactDiff, a large dataset for factual knowledge updates of LLMs with a wide range of scenarios (replacement, archival, new entity insertion, etc.) for a wide range of entities with varying popularity.

It takes the form of a collection of 327K updates reflecting the evolution of knowledge between two instances of the Wikidata knowledge base at two dates, $T_{old}$ and $T_{new}$. As illustrated in Table 1, each triple from the old and new bases is labeled as "new", "obsolete" or "static" to identify the update scenario. These triples are accompanied by elements that enable the application of current update algorithms and the application of domain evaluation metrics, namely the verbalization of triples into natural language form, as well as cloze sentences to test the knowledge of the updated models.

In practice, the release of WikiFactDiff spans the evolution of factual knowledge between $T_{old} =$ *4 January 2021* and $T_{new} =$ *27 February 2023*. The choice of $T_{old}$ is such that new facts in the dataset do not overlap the information in the corpus Pile (Gao et al., 2021), widely used to train many LLMs, e.g., GPT-J (Wang and Komatsuzaki, 2021), GPTNeoX20B (Black et al., 2022), Pythia models (Biderman et al., 2023), and GPT-Neo models (Black et al., 2021). Another strength of the present work is that the creation process fits the time span $[T_{old}, T_{new}]$ of our choice, a property we call **temporal adaptability**. Therefore, new versions of WikiFactDiff may be released to align with the collection dates of datasets other than the Pile dataset.

To illustrate the usability and quality of the corpus, experiments on the knowledge replacement scenario are presented for the main existing factual knowledge update methods. This provides a baseline to the community. Other update types (e.g. archival, new entities) are left for future work since this requires the development of new update algorithms.

The paper is organized as follows: Section 2 introduces the domain and related work; Section 3 gives an overview of the dataset; Section 4 details its creation process; and Section 5 presents the performance of update algorithms on WikiFactDiff.

We publish WikiFactDiff, the source code to build it, and the scripts used to evaluate update algorithms on our dataset, on GitHub and HuggingFace[1].

---

[1]Source code : github.com/Orange-OpenSource/WikiFactDiff, Dataset : huggingface.co/datasets/OrangeInnov/WikiFactDiff

## 2. Related Work

Algorithms and benchmarks have been proposed in the literature to update language models at the level of single facts. Although facts are modeled by triples, the family of algorithms targeted here relies on updating a model based on reading natural language sentences expressing a fact. In the following, we refer to these sentences as **update sentences**. Likewise, the evaluation of these algorithms also relies on presenting verbalized facts to the updated model. Usually, the model is asked to complete cloze test sentences (**cloze tests** for short) with the correct, logically assimilated information. Usually, the update quality is measured on how well the updated LM generalizes on other semantically equivalent but linguistically different cloze tests (this is called **Generalization**). Furthermore, it is tested on other facts to ensure that the update did not impact them negatively (property called **Specificity**). The remainder provides details about these algorithms and benchmarks.

**Knowledge update algorithms.** Si et al. (2023) studied prompting as a way to update GPT3's knowledge, Zhu et al. (2020) proposed fine-tuning specific Transformer blocks to account for knowledge updates; they found that penalizing the parameters being trained when they deviate too much from the original values was essential to avoid altering other facts. We note **FT+L** and **FT** finetuning with and without the previously mentioned penalty respectively. Hypernetworks—neural networks designed to generate or adapt the parameters of another neural network—have also been proposed in De Cao et al. (2021) and Sinitsin et al. (2020). Later, Mitchell et al. (2022) contributed by proposing a fast and scalable hypernetwork approach called **MEND**.

Advances have also been made in locating knowledge in language models: Dai et al. (2022) did experiments with BERT (Devlin et al., 2019) indicating that, given a cloze test for a fact, a small set of neurons, within the mid-level MLP activations, plays a crucial role in accurately predicting the correct answer. Later, Meng et al. (2022) traced the causal effects of hidden state activations within GPT2 (Radford et al., 2018) using causal mediation analysis (Vig et al., 2020) that led to similar results and introduced an update algorithm named **ROME**. Among several algorithms, ROME stood out as the most effective, as it generalized successfully on other cloze tests and had minimal impact on other facts, while other tested algorithms sacrificed one or another. Finally, inspired by ROME, Meng et al. (2023) introduced **MEMIT**, an update algorithm that can perform thousands of updates simultaneously.

**Knowledge update benchmarks.** CounterFact (Meng et al., 2022) and zsRE (Levy et al., 2017) are prominent datasets to evaluate update algorithms. However, their updates are unrealistic because they are generated randomly. It's worth noting that zsRE randomly selected the facts for assessing specificity. Notably, research by Meng et al. (2022) revealed that evaluating specificity on these randomly selected facts is not a sufficiently sensitive metric. In contrast, assessing specificity on neighboring facts was found to be more sensitive and better at highlighting the limitations of update algorithms.

Globally, these datasets contain unrealistic updates. Moreover, all their facts are relations between entities, i.e., they contain no fact relating an entity to a quantity (e.g., the population of Japan being 125.44M), date, or string (e.g. the motto of Stockholm University being "Ex lege libertas"). More importantly, both datasets and update algorithms are limited to replacement updates, while in practice, other relevant update scenarios happen.

## 3. Overview of WikiFactDiff

This section presents basic concepts and terminology regarding triples, before giving a high-level quantitative description of WikiFactDiff and its content, in addition to a comparison between our dataset and CounterFact.

### 3.1. Terminology

One of the popular approaches to organizing facts is to structure them in a knowledge graph, where the nodes are *entities* or *literals* (strings, dates, quantities, etc) and the arcs represent *relations*.

The connection that links the entity $s$ to the entity/literal $o$ with the relation $r$ is represented as a (subject, relation, object) triple or $(s, r, o)$ for short. Thus, the knowledge graph is simply a collection of triples. Sample triples are (France, capital, Paris), (Albert Einstein, date of birth, 14 March 1879), and (Germany, shares borders with, Switzerland).

We call the $(s, r)$-**group** the collection of triples that have $s$ as a subject and $r$ as a relation. We use the term **group** to refer to some collection of triples sharing the same subject and relation. This is a key notion since $(s, r)$-groups usually represent changes that need to be taken together into account to maintain the local consistency of the model's knowledge. For instance, removing the information about Japan's population without injecting the new value does not make sense.

### 3.2. Quantitative description of WikiFactDiff

WikiFactDiff is collected for the time period 4 January 2021 - 27 February 2023 and contains 327K updates. Each update concerns a unique $(s, r)$-group with each of its triples labeled as: **new** when the fact carried by this triple happened after $T_{old}$, i.e., it was not correct before $T_{old}$ and now it is at time $T_{new}$ ; **obsolete** when the fact was correct until $T_{old}$ but it is no more at time $T_{new}$ ; or **static** for facts that did not change between $T_{old}$ and $T_{new}$. This enables the following update scenarios:

- **ReplaceObject**: The updated group contains two triples: one *new*, the other *obsolete*.

- **Archive**: All triples in the group are *obsolete*.

- **AddObject**: The group contains at least one *new* triple and at least one *static* triple, e.g., add a member to a non-empty team.

- **AddRelation**: The group contains only triples *new* and $s$ is not a new entity, which means adding a new relation $r$ to an existing $s$ entity. An example would be adding a "date of death".

- **AddEntity**: When $s$ is a new entity. In this case, all triplets in the group are necessarily labeled *new*.

- **Other**: The rest of the updates. These are other, rarer and more varied situations.

In the case of **AddRelation**, it should be noted that we cannot guarantee that the entity $s$ did not have a relationship $r$ at $T_{old}$ because this information might have been missing in the old Wikidata due to $s$ being a niche entity for example (e.g. the population of an unknown village).

To align with current algorithms, which only handle the scenario of fact replacement, a reduced version of WikiFactDiff is also produced for the purpose of comparing these algorithms, particularly with the CounterFact corpus. This reduced version is called WFD$_{\text{repl}}$. It is constructed by keeping only the groups in WikiFactDiff with exactly one triple labeled "new" and one triple labeled "obsolete". Then, because of their high frequency, we undersample, from this subset, the updates involving the *"population"* relation, by a factor 14, which leaves us with a set called WFD$_{\text{repl}}$ of 10,412 updates. A comparison highlighting the differences between WikiFactDiff and WFD$_{\text{repl}}$ over CounterFact is shown in Table 2.

## 4. Corpus Creation Process

Figure 1 illustrates the creation process of WikiFactDiff. Based on 2 Wikidata raw snapshots at times $T_{old}$ and $T_{new}$, denoted as $\mathbf{W}_{old}$ and $\mathbf{W}_{new}$, the following key stages are processed:

1. **Preprocessing and difference**: Triples in the Wikidata dumps are cleaned and filtered

|  | CounterFact | WFD$_{repl}$ | WikiFactDiff |
|---|---|---|---|
| Triples | 43,838 | 20,824 | 454,365 |
| Subjects | 20,391 | 9,817 | 139,811 |
| Relations | 34 | 157 | 675 |
| Objects | 870 | 12,491 | 111,632 |
| Entity Objects | 870 | 5,578 | 76,015 |
| Literal Objects | 0 | 6,913 | 35,617 |
| Updates | 21,919 | 10,412 | 327,688 |
| ReplaceObject | 21,919 | 10,412 | 32,875 |
| Archive | 0 | 0 | 2,798 |
| AddObject | 0 | 0 | 1,533 |
| AddRelation | 0 | 0 | 155,105 |
| AddEntity | 0 | 0 | 132,857 |
| Other | 0 | 0 | 2,520 |
| Realistic? | ✗ | ✓ | ✓ |
| Temp. Adap.? | ✗ | ✓ | ✓ |

Table 2: Dataset statistics. "Temp. Adap." means "Temporal Adaptability"

to eliminate irrelevant information and metadata, and a naive difference is computed between all $(s,r)$-groups in these two preprocessed snapshots. This results in a partitioning of all triples as $\mathbf{F}^-$, $\mathbf{F}^+$, and $\mathbf{F}^0$ whether they belong to the old or new dump, or both of them.

2. **New entity detection** To type the update scenarios, new entities that appeared during the period $[T_{old}, T_{new}]$ are spotted from $\mathbf{F}^+$.

3. **Classification rules**: All triples from $\mathbf{F}^-$, $\mathbf{F}^+$, and $\mathbf{F}^0$ are screened using hand-crafted rules to categorize them as 'new', 'obsolete', or 'static' (Section 3). This also allows discarding $(s,r)$-groups where the nature of the changes is not entirely clear. This is to ensure that retained factual changes all reflect actual changes in the real world.

4. **Search for Neighboring Facts**: For all triples retained after all the filtering, semantically close triples are identified. This step is crucial for assessing the specificity metrics of update algorithms.

5. **Verbalization**: Finally, update sentences and cloze tests need to be generated to make the factual updates applicable by the algorithm from the literature and evaluate their performance on WikiFactDiff.

As a result, WikiFactDiff includes all the filtered and classified triples, the new entity set, nearest neighbors and verbalizations for each triple. Furthermore, the creation pipeline will be released.

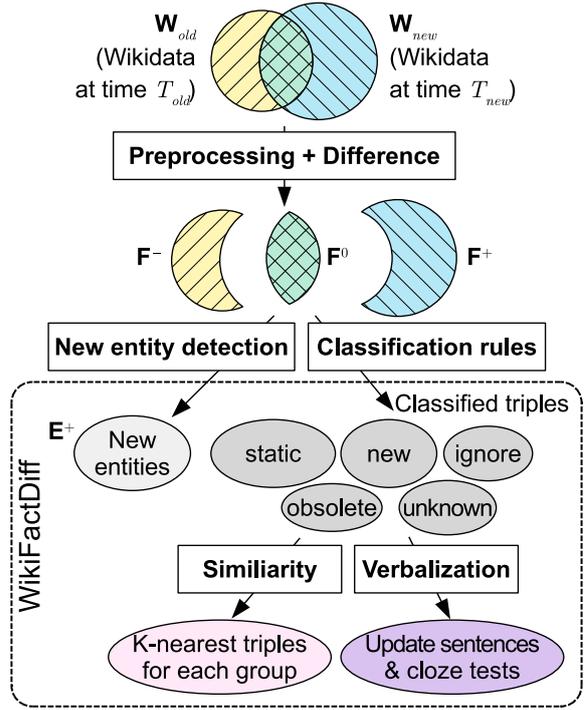

Figure 1: Step-by-step illustration of the creation process of the WikiFactDiff

### 4.1. Preprocessing and Difference Between Wikidata Snapshots

The preprocessing of $\mathbf{W}_{old}$ and $\mathbf{W}_{new}$ consists of several steps, each step filtering out part of the original data.

**Triples vs additional information.** We divide Wikidata snapshots into two parts: the basic facts formalized as triples such as (*Elizabeth II*, *position held*, *Head of the Commonwealth*), and all additional information surrounding those facts, called qualifiers.

In Wikidata, qualifiers allow triples representing simple facts to be expanded on, annotated, or contextualized. For example, the "*start time*" and "*end time*" qualifiers allow to specify the period during which a fact was true. For instance, (*Elizabeth II*, *position held*, *Head of the Commonwealth*) is a Wikidata triple with *start time* and *end time* qualifiers with values "6 February 1952" and "8 September 2022", respectively. Three temporal qualifiers are considered during the creation of the dataset: *start time*, *end time*, and *point in time*.

We only include the basic facts in WikiFactDiff. The additional information, however, is used at some steps of the dataset creation.

**Restricted triples.** Temporal qualifiers are examples of *restrictive qualifiers*[2], i.e., "*qualifiers that restrict or modify the referent of the subject, without which the statement may be inaccurate or*

---
[2] https://www.wikidata.org/wiki/Q61719275

*meaningless"*. For example, the triple (Se7en, *review score*, 83%) is incomplete without the qualification: *reviewed by: Rotten Tomatoes*. We remove all triples with a restrictive qualifier other than *point in time*, *start time*, or *end time*.

**Deprecated triples.** In Wikidata, a rank[3] (*preferred*, *normal*, or *deprecated*) is attached to each triple to assess its relevance. We filter out all triples that have the deprecated rank.

**Meta triples.** Some Wikidata relations have the type *'Wikidata property about Wikimedia entities'*: they are meta relations, employed for the management of Wikimedia projects. Therefore, we remove triples that have a meta-relation.

**Irrelevant or unexploitable object values.** We filter out triples in which the object is a URL or an external ID identifying an entity in another knowledge base. Typically, URLs are links to external websites or various commons media files (e.g. images, videos, documents, audio files, etc.).

Triples whose object are globe coordinates are also filtered out, because they caused discrepancies when computing the difference between the new and old Wikidata: minor deviations arising from floating-point precision often lead to misclassifying equal coordinates as distinct.

Finally, triples whose object is *'some value'* or *'no value'* are also filtered out.

**Irrelevant entities.** Our aim is to keep only triples that are about actual entities in the world. To identify such entities, we rely on Wikipedia: we assume that an entity is relevant only if it has a dedicated Wikipedia article (or rarely, a portion of a Wikipedia article). Moreover, this page must not be a list, category, template, or disambiguation page. Entities that do not meet this condition are filtered out; all triples containing one of those entities are also removed.

**Outdated values in temporal functional relations.** We call a relation $r$ **temporal functional** if for every subject $s$ in the knowledge graph, the $(s, r)$-group can only possess one element when contextualized in a certain temporal point. For instance, the "population" relation is a temporal functional relation, because there can only be one value for the population in a location at a given point in time. Other temporal functional relations are : life expectancy, capital of a country, head of state, etc. From this definition, if a relation is functional, then it is temporal functional.

Filtering step : If the relation $r$ of an $(s, r)$-group is temporal functional with a 'point in time' qualifier, we keep the most up-to-date triple in this group and we remove the rest. In this way, we keep the most up-to-date information for each version of Wikidata. If some triple in the group does not contain a *point in time* qualifier, we keep only the triple with the *'preferred'* rank if it exists.

We associate each entity to a popularity indicator, based on the number of human visits to its Wikipedia article in the months preceding $T_{new}$. The idea is to allow the community to study how algorithms performances vary depending on this indicator. In the dataset, $(s, r)$-groups are sorted in descending order based on the popularity of their subject $s$.

All facts in $\mathbf{W}^P_{old}$ and $\mathbf{W}^P_{new}$ are reliable triples $(s, r, o)$ with optional time information $[t_{start}, t_{end}]$. If $t_{start}$ or $t_{end}$ is not defined, it is mapped to $-\infty$ and $+\infty$, respectively. For facts with punctual information in time $t$ (e.g., populations), the time interval is set to $[t, +\infty]$.

Finally, the intersection and the difference over the sets of triples coming from $\mathbf{W}^P_{old}$ and in $\mathbf{W}^P_{new}$ are computed to produce the complementary sets $\mathbf{F}^-$ (which are only in $\mathbf{W}^P_{old}$), $\mathbf{F}^+$ (which are only in $\mathbf{W}^P_{new}$), and $\mathbf{F}^0$ (which are in both $\mathbf{W}^P_{old}$ and $\mathbf{W}^P_{new}$).

### 4.2. Detection of New Entities

New entities are tangible or intangible objects that did not exist prior to $T_{old}$. Notable examples include *ChatGPT*, *The 2022 Russia-Ukraine War*, *Lilibet of Sussex*, among others. This set could constitute a benchmark for entity insertion in language models. A new entity is any entity $e$ such that: (i) $e$ is only present in $\mathbf{F}^+$ ; (ii) there exists a triple $(e, r, d)$ where $r$ is a relation denoting the creation date[4] of $e$, and $d$ is a date such that $d > T_{old}$. Condition (ii) is needed because some facts may be prior to $T_{old}$ but the fact was missing in $\mathbf{W}_{old}$. If not discarded, knowledge update experiments may be biased because the fact could appear in the training data of the LLM to update.

### 4.3. Classification Rules

All triples from $\mathbf{F}^-$, $\mathbf{F}^0$, and $\mathbf{F}^+$ are classified using manual rules in order to specify their labels. In addition to the labels 'new', 'obsolete', and 'static' defined in Section 3, two more technical labels are introduced:

- 'ignore' applies to facts that are neither correct at time $T_{old}$, nor $T_{new}$. This is typically the case for facts with a validity interval $[t_{start}, t_{end}] \subset [T_{old}, T_{new}]$.

- 'unknown' is a default label, attributed when no other label can be attributed based on our labeling rules.

---

[3] https://www.wikidata.org/wiki/Help:Ranking

[4] In practice, these relations are 'inception', 'date of birth', 'start time', 'time of discovery or invention', 'date of official opening', 'announcement date', 'point in time', and 'publication date'.

| Feature | Description |
|---|---|
| $t_{start}$ | Start time |
| $t_{end}$ | End time |
| $e \in \mathbf{E}^+$ | Is entity $e$ a new entity? |
| $e \notin \mathbf{F}^x$ | Does entity $e$ appear in a triple from $\mathbf{F}^x$? |
| $r$ is death | Is relation $r$ either 'date of death' or 'date of burial or cremation'? |
| $r$ is temporal | Is relation $r$ temporal functional? |
| $n^-, n^0, n^+$ | Number of triples of $(s,r)$-group which are in $\mathbf{F}^-$, $\mathbf{F}^0$, and $\mathbf{F}^+$, respectively |
| $n$ | Total number of triples of the (s, r)-group |

Table 3: Features of the triple $(s, r, o)$

| Condition | Class |
|---|---|
| $s \in \mathbf{E}^+$ | new |
| $s \notin \mathbf{F}^-$ | unknown |
| $r$ is death $\wedge$ $(s,r,o) \in \mathbf{F}^+ \wedge n = 1 \wedge T_{old} < o < T_{new}$ | new |
| $r$ is death $\wedge$ $\neg((s,r,o) \in \mathbf{F}^+ \wedge n = 1 \wedge T_{old} < o < T_{new})$ | unknown |
| $t_{start} > t_{end}$ | unknown |
| $r$ is temporal $\wedge$ $n^- = 1 \wedge n^+ = 1 \wedge n^0 = 0 \wedge (s,r,o) \in \mathbf{F}^+ \wedge T_{old} < t_{start} < T_{new}$ | new |
| $r$ is temporal $\wedge$ $n^- = 1 \wedge n^+ = 1 \wedge n^0 = 0 \wedge (s,r,o) \in \mathbf{F}^+ \wedge T_{old} < t_{start} < T_{new} \wedge (t_{end} = +\infty \vee t_{end} > T_{new})$ | new |
| $(s,r,o) \in \mathbf{F}^+ \wedge T_{old} < t_{start} < T_{new}$ | new |
| $t_{end} < T_{old}$ | ignore |
| $t_{end} = +\infty \wedge t_{start} < T_{old}$ | static |
| $t_{end} = +\infty \wedge T_{old} < t_{start} < T_{new}$ | new |
| $t_{start} > T_{old}$ | ignore |
| $T_{old} < t_{start} < T_{new} \wedge T_{old} < t_{end} < T_{new}$ | ignore |
| $t_{start} < T_{old} \wedge T_{old} < t_{end} < T_{new}$ | obsolete |
| $t_{start} < T_{old} \wedge t_{end} > T_{new}$ | static |
| $T_{old} < t_{end} < T_{new}$ | obsolete |
| $t_{end} > T_{new}$ | static |
| $(s,r,o) \in \mathbf{F}^- \wedge t_{end} < T_{old}$ | ignore |
| $(s,r,o) \in \mathbf{F}^+ \wedge o \in \mathbf{E}^+$ | new |

Table 4: Classification rule list for a given triple $(s, r, o)$. These rules are executed sequentially.

These labels are key pieces of information because their distribution within a given $(s, r)$-group determines the update scenario. For example, a group of 2 triples, one labeled as 'obsolete' and the other labeled as 'new', corresponds to a replacement scenario, similar to the (s, r)-group (USA, head of government) in Table 1.

Table 3 lists the variables and predicates used in the classification rules of each triple $(s, r, o)$, Table 4 reports these rules. For a given triple $(s, r, o)$, the rules are tested in the order they appear in the list. As soon as a rule is evaluated to true, the triple is assigned the corresponding class. If no rule could be applied, the triple remains with the class 'unknown'.

Then, an additional step is performed on every triple of $(s, r)$-groups of size 2 ($n = 2$) where $r$ is a temporal functional relation. Given the pair of triples $(s, r, o_1)$ and $(s, r, o_2)$, if one of them is labeled as 'new' and the other is in $\mathbf{F}^-$, this other one gets assigned the class 'obsolete'.

The final step in processing involves removing anomalies that may have been found in the collected updates. An anomaly occurs when an object that has been tagged shows up multiple times within the same $(s, r)$-group with potentially different labels. For instance, an anomaly would be if an object is simultaneously tagged as "obsolete" and "static" within the same group. Such anomalies are not common; there have only been 2,167 groups in this situation. The method used to address these anomalies are explained in Appendix C.

Upon the completion of this procedure, all $(s, r)$-groups with at least one triple labeled as 'unknown' are discarded to only consider knowledge updates where the change is perfectly understood. Then, all the triples labeled as 'ignore' are removed from the remaining groups. Finally, groups where all triples are labeled with the class 'static' are also filtered. The result is a collection of $(s, r)$ groups where at least one triple is either 'new' or 'oboslete'.

### 4.4. Search for Neighboring Facts

When a fact is updated, the language model probability distribution is altered, which can degrade its accuracy on other facts. This phenomenon is known as **bleedover**. To enable its detection and measurement, WikiFactDiff comes along with neighboring facts that are susceptible to be altered negatively for a given $(s, r)$-group to update. The section explains how this is performed.

It was shown in Meng et al. (2022) that, given an update in a replacement scenario from a fact $(s, r, o)$, the facts with similar relation and object $(s', r, o)$ (with $s' \neq s$) are more susceptible to be impacted compared to random facts (where no significant alteration was reported). For instance, updating (Albert Einstein, specialty, *Physics*) to (Albert Einstein, specialty, *Biology*) can degrade the accuracy of the model on the fact (Isaac Newton, specialty, Physics). The motivation is that because $s$ and $s'$ share the same relation-object pair, their latent representations are close and thus the properties of $s'$ are more susceptible to bleedover. This method cannot work in our setup because it is not guaranteed that some fact $(s', r, o)$ exists for all possible $(s, r, o)$. The reason is that, in WikiFactDiff, the objects in triples are not limited to entities. They can also be literal. For instance, if updating *(Seattle, population, 733.92K)*, finding a close entity to *Seattle* using Meng's method means finding another entity with the exact same population, which is very unlikely to happen. Moreover, even in the case of entity objects, the specificity of the subject can be such that no adequate triple exists. To circumvent this problem, neighboring triples for a triple $(s, r, o)$ are defined as triples $(s', r, o')$ where $s'$ is an entity similar to $s$. Intuitively, this strategy relaxes the constraint on $o$ but strengthens the one on $s$.

Basically, the similarity between two entities is computed as a cosine similarity between TF-IDF vectors representing each entity. Let $\mathbf{W}_{old}^E$ (resp. $\mathbf{W}_{new}^E$) be the set of all triples in $\mathbf{W}_{old}$ (resp. $\mathbf{W}_{new}$) whose object is an entity (not a literal). For each entity $s$, a list of features $I(s)$ is constructed as

$$[s] \oplus [o \mid (s, r, o) \in \mathbf{W}_{old}^E] \oplus [(r, o) \mid (s, r, o) \in \mathbf{W}_{old}^E]$$

where $\oplus$ denotes the concatenation operator on lists. If $s$ is not present in $\mathbf{W}_{old}^E$, $I(s)$ is retrieved from $\mathbf{W}_{new}^E$ instead. Then, TF-IDF sparse representations are computed for all entities $s$ of the dataset, considering each representation $I(s)$ as a document, from which cosine similarities can be derived to compare entities.

For a given triple $(s, r, o)$, the $k$ nearest triples are collected by iterating through the $n$ entities most similar to $s$. One triple maximum of the form $(s', r, o')$ in $\mathbf{W}_{old}$ is selected for each $s'$ in this list, iteratively until reaching $k$ selected triples. In the dataset, the released $k$-nearest-triples have been obtained with $k = 10$ and $n = 500$.

### 4.5. Verbalization

To use existing update algorithms and to evaluate them, update sentences and cloze tests must be generated for all triples in the dataset. For instance, considering the triple *(France, capital, Paris)*, a possible update sentence is *"The capital of France is Paris"*, and cloze tests could be *"The capital of France is ___"*, *"France's official capital is ___"* or *"The capital city of France is none other than ___"*. Note that, since cloze tests are designed for autoregressive models, the blank must be at the end.

Update sentences and cloze tests are generated based on templates of triple verbalizations where both the object and subject are missing, such as *The capital of ___ is ___*. The cloze test for a triple $(s, r, o)$ can be trivially produced from a template by filling the first slot with $s$. Similarly, the update sentence is obtained by injecting $s$ and $o$ into each slot, respectively. Thus, having templates for each relation is sufficient.

The templates are created as follows. First, we randomly sample triples whose subject is one of the 100,000 most popular entities in $\mathbf{W}_{old}$ (Appendix D). Then, for each triple $(s, r, o)$, 10 English verbalizations are generated using ChatGPT[5] (e.g. *"The capital of France is Paris."*). Only verbalizations that (i) contain $s$, and (ii) end with $o$ are kept. Hence, templates for the relation $r$ are obtained by replacing $s$ and $o$ with blanks. However, not all templates are generic enough to be applicable to all triples with the relation $r$. For instance, *"Danish actress ___ was born in ___"* only applies when $s$ is a Danish actress. To filter out such templates, we keep only the 5 most frequent templates for each relation, relying on the idea that templates that apply to all triples with relation $r$ tend to be generated more frequently. Samples of cloze tests built from these templates are shown in Table 5.

---

[5]GPT3.5 version *2023-03-15-preview*

| Subject-relation pair | Cloze test |
|---|---|
| (India, head of state) | India's head of state is ___ |
| (Google, employees) | The number of employees at Google is ___ |
| (Ukraine, BTI Status Index) | The BTI Status Index rated Ukraine at ___ |
| (Lionel Messi, head coach) | Lionel Messi's head coach is ___ |
| (Amazon, chief executive officer) | The CEO of Amazon is ___ |
| (Japan, age of majority) | In Japan, adulthood is recognized at ___ |

Table 5: Samples of cloze tests from WikiFactDiff.

## 5. Experiments

This section evaluates existing atomic knowledge update algorithms on the WFD$_{\text{repl}}$ subset, the restricted version of WikiFactDiff for the sole replacement update scenario (see Sec. 3). The algorithms are ROME, MEMIT, MEND, FT, and FT+L, as implemented by Meng[6]. The model to be updated is GPT-J configured in `bfloat16` precision. These updates are done using one RTX3090 (24GB of VRAM).

An update $u$ from WFD$_{\text{repl}}$ consists in the replacement of a fact $(s, r, o)$ by a fact $(s, r, o^*)$; e.g. (*Japan*, *population*, *125.96M*) by (*Japan*, *population*, *125.44M*). We denote by $c_u + o$ the update sentence used to perform the update on the LM (e.g. *"The population of Japan is 125.44M"*), where $c_u$ refers to the cloze test (*"The population of Japan is ___"*).

After the update is performed, four aspects are evaluated : efficacy, generalization, specificity, and fluency. Efficacy is achieved if the model is able to predict the correct answer on $c_u$. Generalization is achieved if the model can predict the correct answers on alternative cloze tests from a set $C$ (in our experiment, $C$ contains 4 cloze tests). Specificity is achieved by minimizing bleedover, which is a degradation of the capacity to predict the correct answer to cloze tests corresponding to facts that have not been updated. For measuring bleedover, we rely on a set of triples $\{(s_i, r, o_i)\}_i$ selected either randomly[7], or using the K-nearest-triple method. For each triple $(s_i, r, o_i)$, e.g. (*China*, *population*, *1.412B*), we randomly pick a template on $r$ and fill the subject slot with $s_i$ to create a cloze test $c_i$, e.g. *"The population of China is __"*. Finally, *fluency* (Zhang et al., 2018) is the model's ability to produce fluent sentences; it should not decrease after updates. The exact definition of these metrics is available in Table 7. The average performances of each algorithm on WFD$_{\text{repl}}$ are shown in Table 6.

---

[6]github.com/kmeng01/memit

| Algo. | Efficacy-D ↑ | Efficacy-S ↑ | Gen.-D ↑ | Gen.-S ↑ | Bleedover ↓ Random | K-nearest | Fluency ↑ | Time↓ sec/update |
|---|---|---|---|---|---|---|---|---|
| GPT-J | -1.4 ± 0.2 | 44.6 ± 1.0 | -1.3 ± 0.2 | 44.4 ± 0.9 | 0.0 ± 0.0 | 0.0 ± 0.0 | 5.2 ± 0.0 | 0.0 ± 0.0 |
| FT | 45.9 ± 0.5 | 99.6 ± 0.1 | 45.7 ± 0.5 | **99.5 ± 0.1** | 3.3 ± 0.1 | 5.6 ± 0.2 | **0.6 ± 0.0** | 1.4 ± 0.0 |
| FT+L | **12.9 ± 0.6** | **72.9 ± 0.9** | **1.1 ± 0.2** | **53.6 ± 0.8** | 0.1 ± 0.0 | **0.3 ± 0.0** | 5.1 ± 0.0 | 2.4 ± 0.0 |
| MEND | 64.5 ± 0.6 | 99.4 ± 0.1 | 28.8 ± 0.5 | 96.5 ± 0.3 | **0.0 ± 0.0** | 1.0 ± 0.1 | 4.9 ± 0.0 | 1.1 ± 0.0 |
| ROME | **95.5 ± 0.2** | **99.7 ± 0.1** | **59.5 ± 0.6** | 98.0 ± 0.2 | **0.0 ± 0.0** | 0.6 ± 0.1 | **5.2 ± 0.0** | 4.9 ± 0.0 |
| MEMIT‡ | 87.4 ± 0.3 | 99.5 ± 0.1 | 42.1 ± 0.6 | 94.4 ± 0.3 | **0.0 ± 0.0** | **0.3 ± 0.0** | **5.2 ± 0.0** | 41.4 ± 0.2 |
| PROMPT | 58.6 ± 0.5 | 98.9 ± 0.2 | 30.8 ± 0.5 | 93.3 ± 0.4 | 1.1 ± 0.0 | **0.3 ± 0.0** | 4.4 ± 0.0 | **0.0 ± 0.0** |

Table 6: Numerical results of update algorithms on WFD$_{\text{repl}}$ with their respective 95% confidence intervals. **Green underlined** values represent columnwise maxima and **red** values are clear failure of an algorithm on one metric. ‡ indicates algorithms designed for batched updates.

In addition to the update methods mentioned in Section 2, we evaluate the PROMPT algorithm, which consists in influencing the model's knowledge at inference time by prefixing each prompt with $c_u + o^*$. For instance, if we want to update the president of USA to *Joe Biden*, we prefix the model with *"The president of USA is Joe Biden."*. This presents an opportunity to compare two categories of methods: those that modify the model's parameters (as discussed earlier) and those that inject knowledge through prompting. The latter approach, extensively employed in Retrieval-Augmented Generation (RAG) methods (Lewis et al., 2020), sidesteps the challenge of knowledge update by directly infusing the required knowledge into the prompt.

### 5.1. General Results

Our results mainly agree with the ones produced with CounterFact, which indicates that update algorithms perform equally well whether the update is realistic or not : FT generalizes well but fails completely to maintain specificity and fluency; on the other hand, FT+L does not cause bleedover but fails to generalize. Although ROME is the best algorithm overall, the gap with MEND is not as pronounced as in CounterFact, especially regard-ing specificity and generalization. Note also that FT+L's efficacy is not as high as in CounterFact. Finally, PROMPT is competitive with the state of the art on all metrics, except bleedover on random neighbors (this specificity is commented further in Sec. 5.2). Since prompting facts unrelated to the task is rarely done in practice, the high bleedover of PROMPT on random facts is not a critical weakness of the method. However, one should keep in mind that inserting knowledge in an LLM with prompts is limited by the context size, contrary to methods that update the model parameters.

---
[7]Random facts are sampled uniformly from the union of neighbor facts of all instances in WFD$_{\text{repl}}$

| **Efficacy Difference** |
|---|
| $\mathbb{P}^*[o^*\|c_u] - \mathbb{P}^*[o\|c_u]$ |
| **Efficacy Success** |
| $\mathbb{1}_{\mathbb{P}^*[o^*\|c_u] > \mathbb{P}^*[o\|c_u]}$ |
| **Generalization Difference** |
| $\frac{1}{\|C\|} \sum_{c \in C} \mathbb{P}^*[o^*\|c] - \mathbb{P}^*[o\|c]$ |
| **Generalization Success** |
| $\frac{1}{\|C\|} \sum_{c \in C} \mathbb{1}_{\mathbb{P}^*[o^*\|c] > \mathbb{P}^*[o\|c]}$ |
| **Bleedover** |
| $-\frac{1}{\|N\|} \sum_{(c',o') \in N} min(\mathbb{P}^*[o'\|c'] - \mathbb{P}[o'\|c'], 0)$ |
| **Fluency** |
| $\frac{1}{\|C\|} \sum_{c \in C} \frac{2}{3} H_2(G(c)) + \frac{4}{3} H_3(G(c))$ |

Table 7: Metrics for a single update. $\mathbb{P}$ and $\mathbb{P}^*$ are the language model probability function before and after the update, respectively. $H_n(x)$ is the $n$-gram entropy on a text $x$ and $G(c)$ is the model's greedy text generation function given the prompt $c$ (the prompt is not included in $G(c)$).

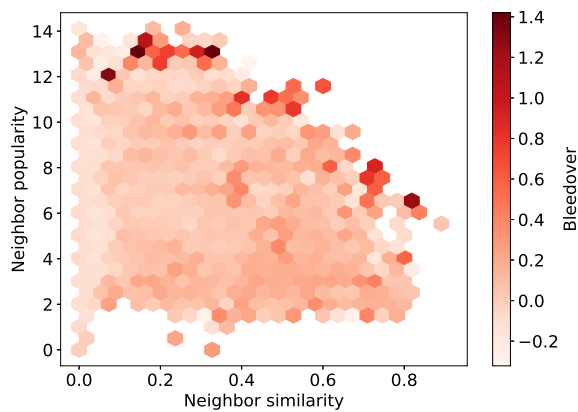

Figure 2: Average impact of neighbor popularity and similarity on bleedover. This metric is normalized for each algorithm to mitigate the variance in bleedover between them.

### 5.2. Effectiveness of K-nearest-triples for Bleedover Detection

For all methods except PROMPT, we observe that bleedover scores are higher on K-nearest-triples than on random neighbors. PROMPT's higher bleedover scores on random neighbors might be explained as follows: prompting a completely unrelated fact before the cloze test creates an unusual setting (e.g. "My Mister has a duration of 90 min. Langenbach has a population of ___") does not appear often in the model's training set which disrupts its behavior.

The fact that all methods relying on LLM parameters update have a higher bleedover score on K-nearest neighbors confirms the relevance of the K-nearest-triples approach. However, the fact that PROMPT is prone to bleedover on random neighbors suggests that bleedover on random facts might be a useful complementary measure.

Finally, there are simple possible improvements to our triple selection method for bleedover detection. Indeed, LMs' knowledge is biased towards popular subjects (Kandpal et al., 2023), therefore we can suspect that the popularity of an entity has some influence on the bleedover probability. We measured this influence by plotting the average bleedover on triples, depending on the popularity of their subject and on their similarity to the updated triple (Figure 2). It appears that both factors have a positive influence on the bleedover probability. For future research, our neighbor selection method for bleedover detection could thus be improved by integrating the subject's popularity.

### 6. Conclusion

In this paper, we introduced WikiFactDiff, a new large knowledge update dataset containing factual knowledge changes over a time period. It considerably extends the range of considered facts and update scenarios previously proposed in the literature (literals ; realistic changes ; e.g., entity insertion and archival ; etc.). As proven by the experiments, the dataset comes with all the necessary material to run and evaluate update algorithms. Moreover, the corpus creation process is adaptable to new periods. The main direction for future work is now to design algorithms able to deal with the challenging new proposed update scenarios.

### 7. Bibliographical References

### A. How are temporal functional relations identified?

Temporal functional relations are identified using the "property constraint" section of a relation in Wikidata.

If a relation contains a property constraint that is either "single-value constraint" or "single-best-value constraint", then the relation is functional. If, in addition to that, this property constraint has a qualifier "separator" that contains "start time", "end time", or "point in time", then this relation is temporal functional.

For example, "head of state" is a temporal functional relation (www.wikidata.org/wiki/Property:P35)

### B. Prompt used to generate cloze tests with ChatGPT

Here is the system prompt (also called pre-prompt) of ChatGPT that is used to generate cloze tests:

```
You are an advanced knowledge triple
verbalization system.
You take as input a knowledge triple
(subject, relation, object) and generate
a list of 10 linguistically diverse
verbalizations of the triple.
For example, the input could be : (France,
capital, Paris) and one of your
verbalizations may be : "The capital of
France is Paris".

The veracity of the knowledge triple does
not affect the quality of your generation.

Examples of correct verbalizations:
- (Matriak, instance of, university) -->
"Matriak is a university."
- (Johnathan Smith, date of death, 11-05-
2012) --> "Johnathan Smith died in 11-05-
2012."
- (Tranquility Base Hotel & Casino,
follows, AM) --> "Tranquility Base Hotel
& Casino follows AM."
- (Paris, named after, Parisii) --> "Paris
was named after Parisii."
```

And here is the main prompt:

```
Here is the knowledge triple to
verbalize: ([SUB], [REL], [OBJ]). Your
sentences should be concise and end
with the term [OBJ].

Due to the ambiguity that could arise
from the provided labels, here is
their meaning:
- (subject) "[SUB]" : "[SUB_DEF]"
- (relation) "[REL]" : "[REL_DEF]"
- (object) "[OBJ]" : "[OBJ_DEF]"

Finally, here is an example where
the relation "[REL]" is employed :
([EXP_SUB], [REL], [EXP_OBJ]).
```

We used greedy search with temperature = 0, no frequency penalty, no presence penalty, and with a maximum number of generated tokens equal to 800.

### C. Addressing anomalies

Let $l_1, l_2, \ldots, l_k$ be all the labels associated to the object $o$ in a group and suppose there is anomaly, meaning $k > 1$. Here is the case-by-case treatment for encoutered anomalies:

- $\forall i, l_i =$ **obsolete** : One object labeled **obsolete** is kept and the rest is deleted.

- $\forall i, l_i \in \{$**new**, **static**$\}$ : One object labeled **static**, or as a second choice **new** if **static** does not exist, is kept and the rest is deleted.

- **Else** : The whole group is deleted.

1435 groups were deleted by applying this treatment.

### D. Verbalization : Triple sampling process

The triples to verbalize are selected as follows:

- First, the 100,000 most popular entities (using our popularity measure) are shuffled

- For each of these entities noted $e$, we select all the $(e, r)$-groups. Only the first triple in each group is selected for verbalization by ChatGPT.

- A relation is not selected for verbalization once it reaches 100 verbalizations.

We verbalized 26058 triples from $\mathbf{W}_{old}$ during this process.